# iASiS: Towards Heterogeneous Big Data Analysis for Personalized Medicine

Anastasia Krithara[*], Fotis Aisopos[*], Vassiliki Rentoumi[*], Anastasios Nentidis[*], Konstantinos Bougatiotis[*], Maria-Esther Vidal[#], Ernestina Menasalvas[+], Alejandro Rodriguez-Gonzalez[+], Eleftherios G. Samaras^, Peter Garrard^, Maria Torrente[$], Mariano Provencio Pulla[$], Nikos Dimakopoulos[-], Rui Mauricio[%], Jordi Rambla De Argila[~], Gian Gaetano Tartaglia[~] and George Paliouras[*]

[*] *Institute of Informatics and Telecommunications, NCSR "Demokritos"*
*Patriarchou Grigoriou E' & 27 Neapoleos St., Aghia Paraskevi, Athens, Greece*
*{ akrithara, fotis.aisopos, vassiliki.rentoumi, tasosnent, bogas.ko, paliourg }@iit.demokritos.gr*

[#] *Gottfried Wilhelm Leibniz Universität Hannover*
*Welfengarten 1, Hannover, Germany*
*maria.vidal@tib.eu*

[+] *Department of Computer Systems Languages and Sw Engeneering, Universidad Politécnica de Madrid*
*Centro de Tecnología Biomédica (CTB), Campus de Montegancedo s/n, Pozuelo de Alarcón, Madrid, Spain*
*{ ernestina.menasalvas, alejandro.rg }@upm.es*

[^] *St George's University of London*
*Cranmer Terrace, London SW17 0RE, United Kingdom*
*{ esamaras, pgarrard }@sgul.ac.uk*

[$] *Medical Oncology Department, Puerta de Hierro University Hospital, Servicio Madrileño de Salud*
*Calle Manuel de Falla, 1, Majadahonda, Madrid, Spain*
*{ mtorrente80, mprovenciop }@gmail.com*

[-] *Innovation Lab, Athens Technology Center*
*Rizariou 10, Chalandri, Athens, Greece*
*n.dimakopoulos@atc.gr*

[%] *Alzheimer's Research UK*
*3 Riverside, Granta Park, Cambridge, United Kingdom*
*rui.mauricio@alzheimersresearchuk.org*

[~] *Centre for Genomic Regulation*
*C/ Dr. Aiguader, 88, PRBB Building, Barcelona, Spain*
*{ jordi.rambla, gian.tartaglia }@crg.eu*

***Abstract*— The vision of IASIS project is to turn the wave of big biomedical data heading our way into actionable knowledge for decision makers. This is achieved by integrating data from disparate sources, including genomics, electronic health records and bibliography, and applying advanced analytics methods to discover useful patterns. The goal is to turn large amounts of available data into actionable information to authorities for planning public health activities and policies. The integration and analysis of these heterogeneous sources of information will enable the best decisions to be made, allowing for diagnosis and treatment to be personalised to each individual. The project offers a common representation schema for the heterogeneous data sources. The iASiS infrastructure is able to convert clinical notes into usable data, combine them with genomic data, related bibliography, image data and more, and create a global knowledge base. This facilitates the use of intelligent methods in order to discover useful patterns across different resources. Using semantic integration of data gives the opportunity to generate information that is rich, auditable and reliable. This information can be used to provide better care, reduce errors and create more confidence in sharing data, thus providing more insights and opportunities. Data resources for two different disease categories are explored within the iASiS use cases, dementia and lung cancer.**

## I. Introduction

The use of big data in healthcare is in its early days, and most of the potential for value creation remains unclaimed. Towards this direction, iASiS aims to enable comprehensive access to data from disparate sources and results of analysis, in order to produce actionable knowledge for policy-making, within the domain of personalised medicine [1]. The project develops a platform that collects, integrates, and analyses big data from disparate sources, providing useful insights and high-level analysis on an aggregated knowledge graph.

Turning these large amounts of data into actionable knowledge will enable better planning of public health activities and policies. In particular, we envision the use of such knowledge for improved diagnosis and treatment, personalised to each individual. In this context, the project aims at homogenising and combining various available datasets to perform a thorough analysis that will identify patterns to be assessed by clinicians and policy makers. This tool will support the aforementioned users in decision making with a set of statistics, graphs, patterns and personalised recommendations based on the characteristics of each patient, to improve the provisioning of public health services.

Given the above, the specific objectives of iASiS are:

a) to design a unified conceptual schema to represent all the diverse sources of available data,

b) to build an adaptive system able to manage data and content collected incrementally,

c) to provide actionable knowledge about disease diagnosis, prognosis, and treatment to decision makers,

d) to promote cooperation among clinicians and policy makers, and

e) to define solid privacy- and trust-aware strategies for the use of data and the discovered knowledge.

## II. Related Work

Currently there are several efforts made by various international projects, which try to process and integrate various types of data towards the achievement of precision medicine. In specific Dementias Platform UK (DPUK)[1] is a public-private partnership funded by the Medical Research Council. DPUK brings data from multiple cohorts into the DPUK Data Portal. By joining data DPUK provides an integrated and collaborative environment, bringing together scientists from academia and industry to share knowledge and conduct joint research programs with the goal to fight to develop effective treatments for Dementia fast. The large number of individuals DPUK cohorts allows key research questions to be answered more rigorously and more rapidly than would otherwise be possible. In essence, DPUK involves the collection of various dementia related data sources ranging from HER (electronic health records), brain imaging and brain cell data. The DPUK platform encourages the development of new tools and resources to deliver research to accelerate pathways for future medicines. Moreover, TRACERx [2] (TRAcking Cancer Evolution through therapy (Rx)) is a translational research study aimed at transforming our understanding of cancer evolution and take a practical step towards an era of precision medicine. It employs observational cohort data and aims at analyzing intra-tumor heterogeneity in order to help the development of novel, targeted and immune based therapies.

Both projects described in the previous section tackle one disease each, Dementia and Cancer respectively. On the other hand, iASiS tackles both Lung Cancer and Alzheimer's diseases. What is more, DPUK employs only observational data and TRACERx focuses on analyzing genomic data. On the other hand, iASiS moves a step forward as it combines both observational, genomic but also Open Data in order to combine and integrate them into a unified Knowledge Graph.

Furthermore, the Horizon2020 project MIDAS [3] (Meaningful Integration of Data Analytics and Services) aims at the integration of various heterogeneous health care data into a unified data platform which will help policy makers to make informed decisions. MIDAS project is very relevant to IASIS, nevertheless it mainly targets policy makers as end-users, while iASiS targets both clinicians and policy makers. It also involves various epidemiology challenges rather than providing insights for specific diseases as iASiS. OHDSI [4] (Observational Health Data and Informatics) is a multi-stakeholder, interdisciplinary collaborative which employs observational health data to perform large-scale analytics for precision medicine. While OHDSI focuses only on observational data, iASiS employs a multitude of data including also observational.

Lastly, OpenPHACTS [5] creates an information infrastructure for applied life science and drug discovery research and development. OpenPHACTS integrates data sources from various data bases in order to provide relationships of semantically linked life science data. OpenPHACTS mainly integrates structured data bases, while iASiS employs all kind of Open Data (literature, structured and unstructured data) along with other various types of information under the same unified Knowledge Graph.

## III. Methodology

The methodological approach followed by iASiS towards personalized medicine is based on three main pillars:

A. Aggregating multi-modal big datasets from various sources, such as clinical records, genomics, computed tomography (CT) scans etc., and harmonizing those in a unified schema representation

B. A first level analysis of the aforementioned data, using Natural Language Processing (NLP) or other machine learning techniques to extract meaningful knowledge in a structured form

C. High level analysis models, applying link prediction or community detection to identify meaningful patterns on patients and compounds and provide predictive analytics

### A. *Data Acquisition and Unified Representation*

A major task towards the defined objectives, is selecting the available data sources, defining pilot plans and collecting end-user requirements for the two project use cases (Dementia [2] and Lung Cancer [3]).

In specific, data exploited in iASiS include:

- Electronic Health Records (EHRs) and medical images from patients. Access to clinical data for lung cancer was provided by the hospital Puerta de Hierro de Madrid. Respectively dementia-related datasets have been made available to the consortium by UK CRIS (Clinical Record Interactive Search) [4] and the UK Biobank [5].

- Genomic data. Acquired from various sources, such as the European Genome-Phenome Archive [6], NCBI ClinVar [7] and COSMIC [8].

- Open literature data from PubMed and open structured data from various databases and ontologies (e.g. Drugbank [9] and UMLS [10]).

Specialized tools were built for harvesting, semantic indexing and performing an initial analysis of all these datasets [11]. The focus was on establishing interoperability across the different datasets.

---

[1] https://www.dementiasplatform.uk/
[2] http://tracerx.co.uk/
[3] http://www.midasproject.eu/about/
[4] https://www.ohdsi.org/
[5] https://www.openphacts.org/

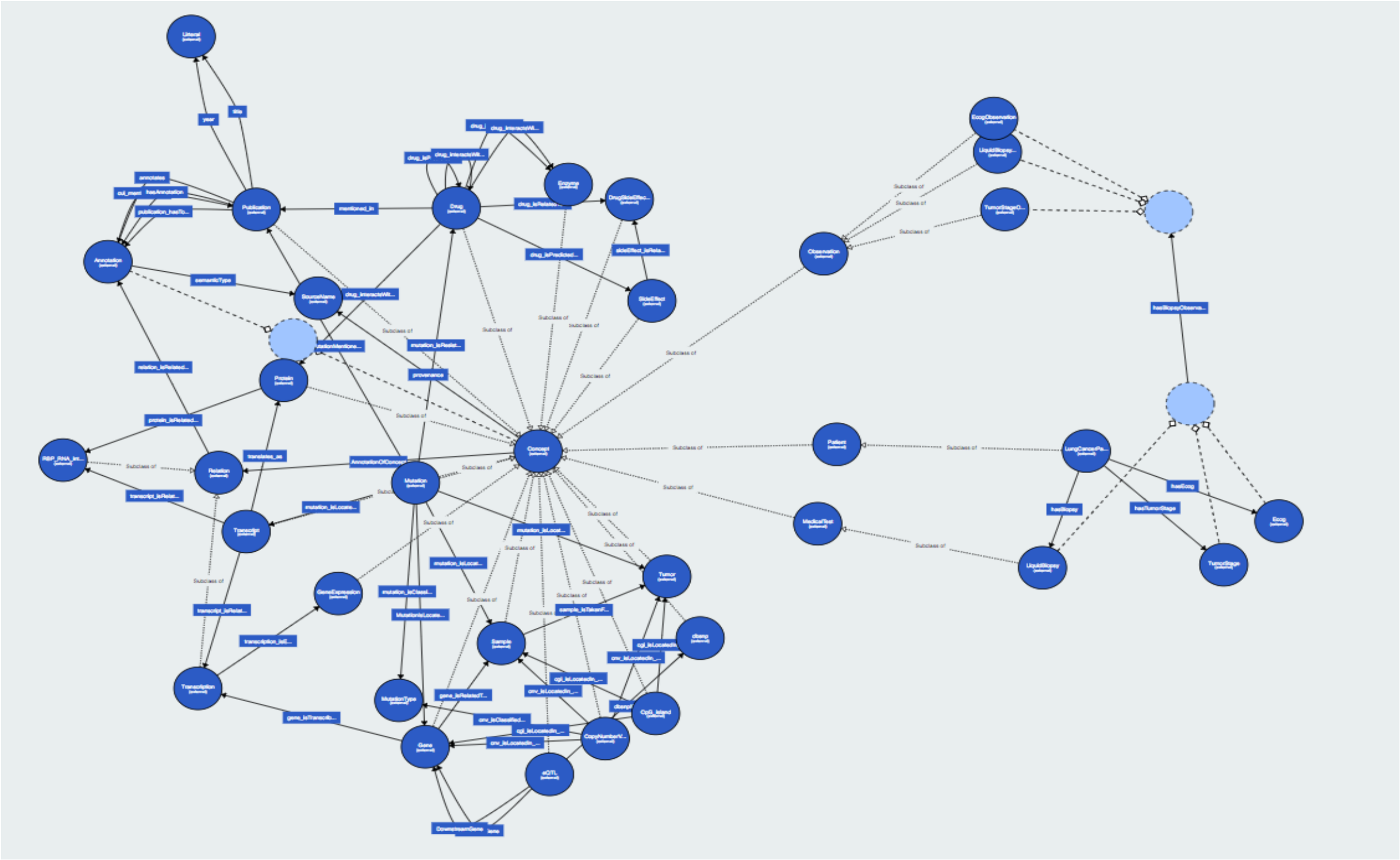

**Figure 1: Visualisation of the iASiS Unified Schema**

Critical to the achievement of interoperability was the definition and implementation of a unified schema, in the form of the iASiS Knowledge Graph (KG) [12]. Knowledge Graphs present a very popular approach for aggregating biomedical big data in a large-scale knowledge base in the form of triples, illustrating biomedical entity pairs associated via a specific relation. The creation of Knowledge Graphs allows for a subsequent mining and analysis, in order to identify useful patterns and meta-associations, as aimed by iASiS. An overview of the iASiS KG can be seen in Figure 1.

### *B. Data Semantic Indexing and First Level Analysis*

As mentioned, iASiS goes beyond the state-of-the-art by leveraging open access information and resources and employing powerful modeling of medical phenomena, in order to test research hypotheses, evaluate new treatments, predict outcomes and propose practices and, overall, empower well-documented personalised medicine.

As a first step, the integration of data from various sources (Electronic Health Records, literature, open structured and genomics sources) has been described in the previous section. To this end, a unified schema has been defined and the iASiS Knowledge Graph has been implemented, which semantically connects all available knowledge. To achieve the goal of fine grained semantic indexing, innovative methods and technologies [15] have been applied to different types of data .

Moreover, NLP techniques have been applied to extract rich knowledge encoded in free text in Electronic Health Records [16], in order to integrate the results into the iASiS Knowledge Graph. Those techniques reconstruct the medical history of each patient, with the use of semantic annotators for entity recognition. However, retrieved patient medical records also include image files, such as CT scans. To analyse those, an innovative module extracting semantic (2D and 3D) and agnostic features (deep features) from CT images has been implemented and applied to an open access image database. The extracted features are used in a predictive modelling process, using Convolutional Neural Network models to search for patterns that support the discrimination of malignant and non-malignant nodules [17].

Concerning genomics, by combining large-scale data on genetic variants which affect the expression of distant genes (“trans- eQTLs”), with information on protein-RNA interactions and clinically relevant genomic variation [18], several candidate molecular interactions of interest have been identified that may have impact to diseases studied in the project [19].

Lastly, concerning open datasets, state-of-the-art text mining and machine learning techniques have been extended, in order to analyse biomedical literature and combine it with knowledge from relevant structured databases. Those techniques provide an Open Data sub-graph, illustrated semantified entities and relations identified in the open data sources. iASiS has also developed an innovative Patients’

Textual Data Analysis module that deploys sophisticated feature extraction methodologies and machine learning techniques, to provide accurate risk assessment regarding the Alzheimer's disease for patients that have participated in a specific cognitive awareness task [20].

### C. *Knowledge and Pattern Extraction*

By bringing different sources of data together, as well as the aforementioned first level analysis results, the Knowledge Graph lends itself to further, high-level analysis. Such analysis has been performed, aiming at discovering latent causal relations between biomedical entities of different sources and developing powerful medical inference tools [13]. This allows for better modeling of complex relations for the diseases under question.

On top of the iASiS Knowledge Graph (KG), mining tools that extract knowledge and uncover unknown patterns from the combination of the aforementioned data have been developed. These mining techniques extend existing community detection approaches, to exploit semantics encoded in the KG, while scalability and efficiency are enforced. Thus, they create communities of biomedical entities, e.g., drugs, genes, or phenotypes, which are similar in terms of domain-specific similarity measures, as well as strongly related based on functional units in the biomedical domain.

Integrated data and results of the aforementioned analysis and identified patterns are provided to decision makers via the iASiS platform software. The iASiS platform identifies and visualises patterns regarding patient demographics and associated entities (diseases, drugs, side effects etc.), as well as patient survival curves. Also it provides predictive tasks on biomedical entities and relations (such as drug interactions or side-effects), based on the combination of heterogeneous knowledge.

## IV. Results

### A. *First Platform Prototype and User Interface*

The first prototype of the iASiS platform has been designed and implemented, including an adaptive Graphical User Interface that can manage aggregated data and analysis results. Thus, the platform has a dual scope: First, it provides user-friendly access to collected, integrated and analysed data. Additionally, it provides it links data to outcomes and analysis results. Based on this information, clinicians and policy makers can efficiently and transparently make decisions that are tailored to individuals, contributing to the achievement of personalised medicine. Figure 2 provides a view of the platform interface, by illustrating an example of pattern extraction for a specific subset of the population.

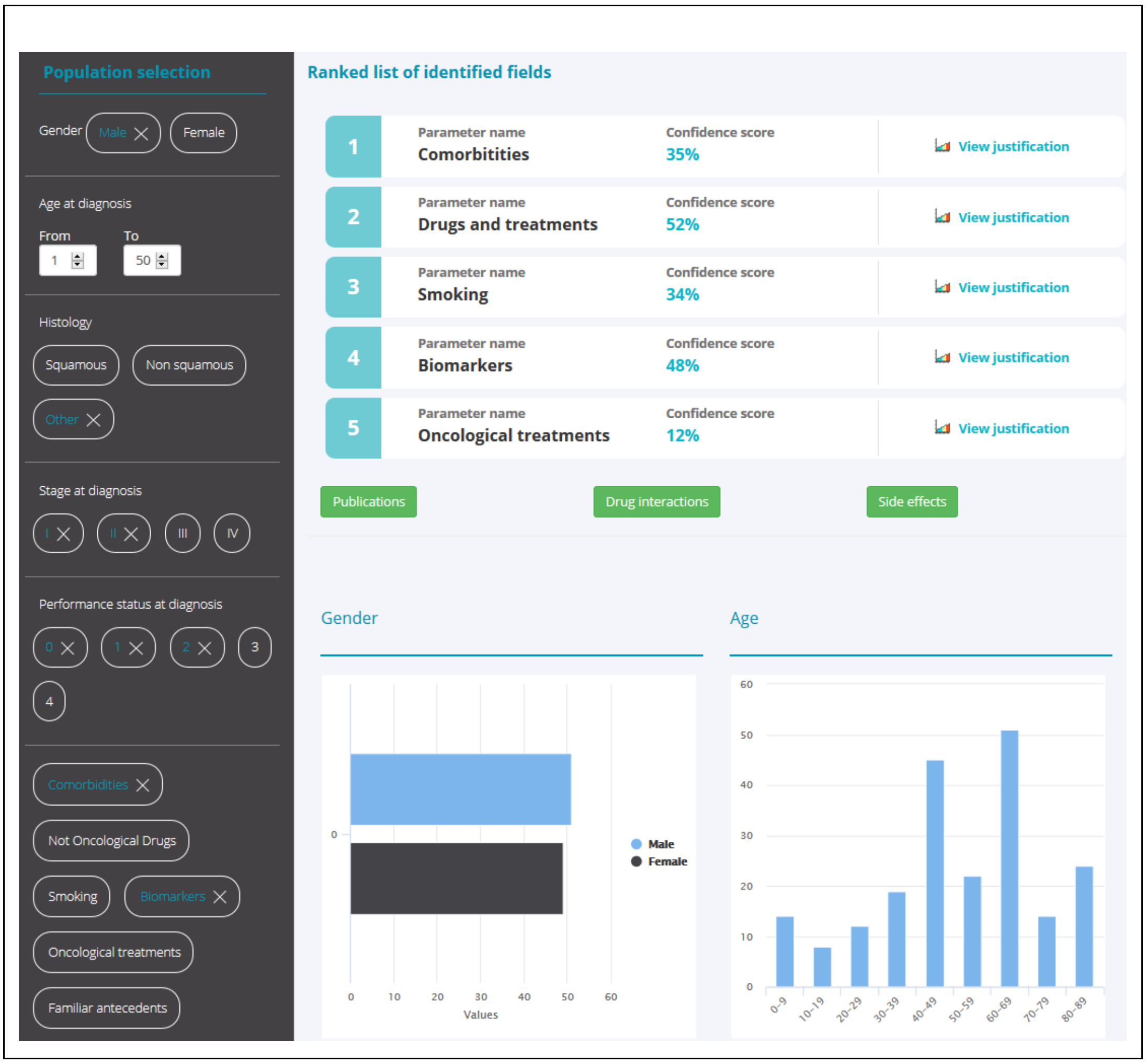

**Figure 2: Example of pattern extraction through Platform user interface**

## B. *Addressing Real User Needs*

The usability and usefulness of the iASiS platform is guaranteed by the continuous involvement of users in all stages of the development process. This process started in the first months of the project, with the identification of user requirements. The process of deriving the user requirements is triggered through illustrative schematic scenarios that have been developed for both pilots, highlighting desirable features and interactions.

As an example, one of the scenarios developed for the lung cancer pilot includes the identification of patterns in the data of long-surviving lung cancer patients [14]. Such patterns may capture associations of family history, treatments, response to treatments, toxicities, comorbidities and molecular mechanisms, with the patient's outcome. The discovery of such patterns can help decision makers in prognosing the progress of disease at a personalised level and even identifying ways to improve the prognosis for specific groups of patients. The iASiS platform identifies long-surviving lung cancer patients, analyses key factors that may affect survival, and compares long-surviving patients with the rest of the patients, in order to identify potential causalities.

Similarly, an example scenario for the dementia pilot, studies the relation of symptomatic treatments within a given class of drugs with different patient types, based mainly on patient's genetic (allelic) status. The iASiS platform identifies alleles of genes associated with an elevated risk of Alzheimer's Disease, based on current bibliography, drug trials and related current policies.

# V. SUMMARY

## A. *Expected results until the end of the Project*

By the end of the project, the consortium is planning to integrate knowledge from more biomedical datasets and ontologies, as well as from all the Electronic Health Records of the participating hospitals (including data from UK-CRIS, Optima and UK-Biobank datasets). All individual modules will be extended to reassure access to up-to-date knowledge extracted from the different sources, and consequently lead to a richer Knowledge Graph pertaining to the Lung Cancer and Dementia use cases. The platform second prototype, will incorporate changes, based on the user evaluation of the first prototype. Additionally, it will integrate more knowledge from the selected data sources and will provide more functionalities, which are expected to lead to more concrete insights for personalised diagnosis and treatment.

## B. *Potential Impacts*

iASiS aims at a significant impact on the EU healthcare system, ICT industry, and generally the wider society.

The project pursues the goal of turning large amounts of data into actionable information for decision makers. Thus the iASiS platform can provide an important tool supporting patients' treatment, providing useful knowledge to the medical professionals and policy makers, such as statistics, survival curves or expected drug interactions, in accordance to patients' habits and personal characteristics. This will enable the best decisions to be made, allowing for diagnosis and treatment to be personalised to each individual.

## C. *Data Privacy*

Detailed data management plans for both the Dementia and Lung Cancer use cases have been documented. These plans describe how data processing in iASiS will respect the policies associated with each data source, adhering also to the EU data regulations. This process is greatly assisted by the Ethics Committee of the project, which was established in the early stages. The Committee is led by an experienced external expert and it reviews the pertinent procedures, permissions and documents together with the privacy and trust-aware strategies.

## D. *Outreach*

Beyond the scientific and technological progress described, the project contributors have worked on reaching out to the communities of big data and personalised medicine, as well as the medical communities in the two diseases addressed in the project. An Advisory Board was established, comprising policy makers and clinicians with significant expertise and influence.

# ACKNOWLEDGMENT

This paper is supported by European Union's Horizon 2020 research and innovation programme under grant agreement No. 727658, project IASIS (Integration and analysis of heterogeneous big data for precision medicine and suggested treatments for different types of patients).